\documentclass[10pt,twocolumn,letterpaper]{article}

\usepackage{cvpr}
\usepackage{times}
\usepackage{epsfig}
\usepackage{graphicx}
\usepackage{amsmath}
\usepackage{amssymb}
\usepackage{multirow}
\usepackage{subcaption}

\usepackage[pagebackref=true,breaklinks=true,letterpaper=true,colorlinks,bookmarks=false]{hyperref}

\cvprfinalcopy 

\def\cvprPaperID{792} 

\ifcvprfinal\fi
\begin{document}
\title{Dense Captioning with Joint Inference and Visual Context}
\author{Linjie Yang~~~~~~~~~~~Kevin Tang~~~~~~~~~~~~Jianchao Yang~~~~~~~~~~Li-Jia Li\\
Snap Inc.\\
\tt\small\{linjie.yang, kevin.tang, jianchao.yang\}@snap.com~~~~~lijiali@cs.stanford.edu\\
}

\maketitle

\begin{abstract}
Dense captioning is a newly emerging computer vision topic for understanding images with dense language descriptions. The goal is to densely detect visual concepts (e.g., objects, object parts, and interactions between them) from images, labeling each with a short descriptive phrase. We identify two key challenges of dense captioning that need to be properly addressed when tackling the problem. First, dense visual concept annotations in each image are associated with highly overlapping target regions, making accurate localization of each visual concept challenging. Second, the large amount of visual concepts makes it hard to recognize each of them by appearance alone. We propose a new model pipeline based on two novel ideas, joint inference and context fusion, to alleviate these two challenges. We design our model architecture in a methodical manner and thoroughly evaluate the variations in architecture. Our final model,  compact and efficient, achieves state-of-the-art accuracy on Visual Genome~\cite{Krishna2016visual} for dense captioning with a relative gain of 73\% compared to the previous best algorithm. Qualitative experiments also reveal the semantic capabilities of our model in dense captioning. Our code is released at \url{https://github.com/linjieyangsc/densecap}.
%
%
%
%
%
%
\vspace{-4pt}
\end{abstract}

\section{Introduction}

\begin{figure}[t]\centering
\includegraphics[width=1.0\linewidth]{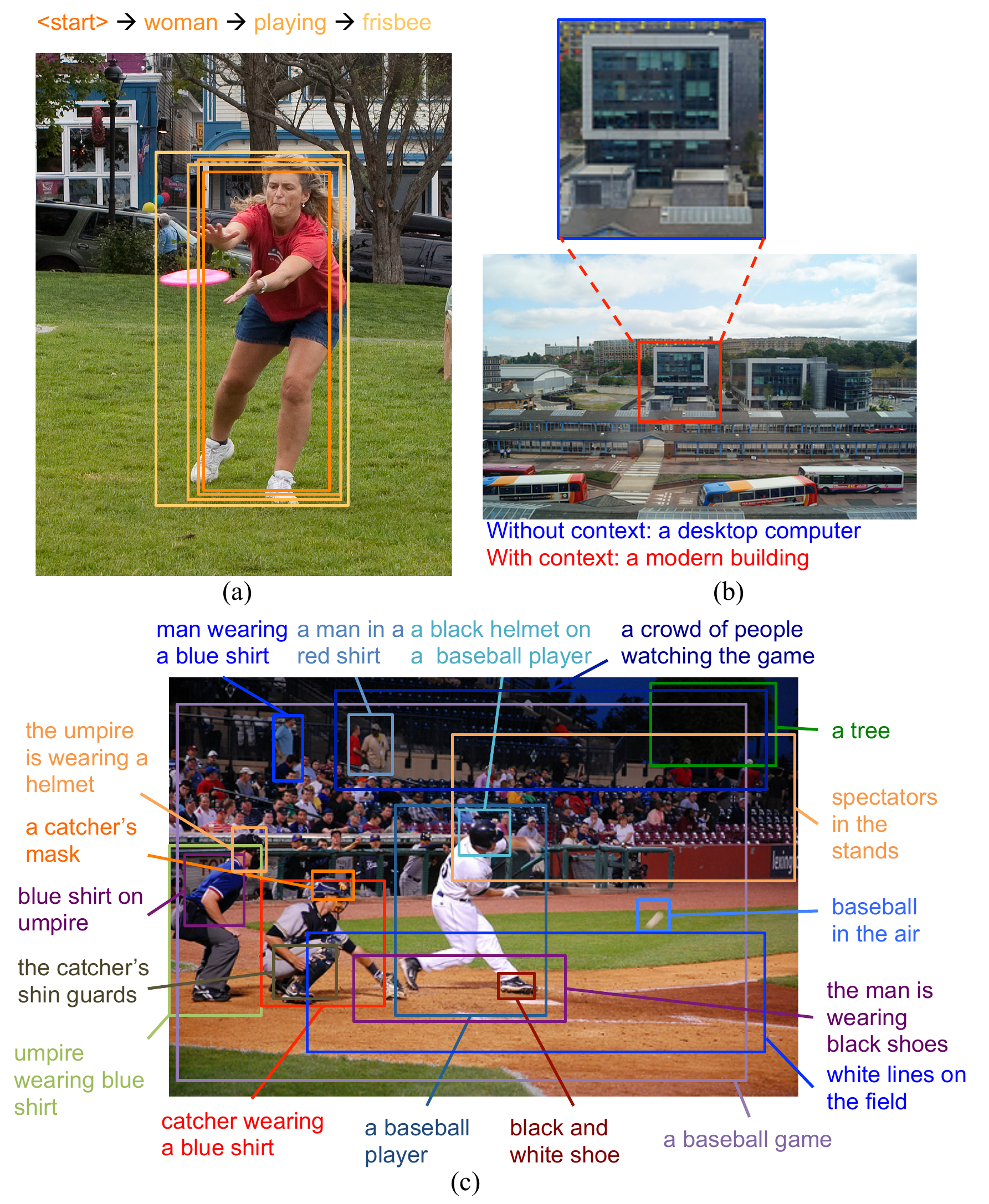}
\caption{Illustration of our approach for dense captioning. (a) For a region proposal, the bounding box can adapt and improve with the caption word by word. In this example, the bounding box is guided by the caption to include the frisbee, even though the initial position was ambiguous. (b) The object in the red box is hard to recognize as a building without the context of the whole image. (c) An example image overlaid with the most confident region descriptions by our model. }
%
%
%
%
%
%
\label{fig:overview}
\vspace{-7pt}
\end{figure}

The computer vision community has recently witnessed the success of deep neural networks for image captioning, in which a sentence is generated to describe a given image. Challenging as it seems, a list of pioneering approaches~\cite{Donahue2015long}~\cite{Fang2015captions}~\cite{Vinyals2015show}~\cite{Xu2015show} have achieved remarkable success on datasets such as Flicker30k~\cite{Young2014image} and MS COCO~\cite{Chen2015coco}. For evaluation, metrics in natural language processing are employed to measure the similarity between ground truth captions and predictions, such as BLEU~\cite{Papineni2002bleu}, Meteor~\cite{Banerjee2005meteor}, and CIDEr~\cite{Vedantam2015cider}. However, the holistic image descriptions from these datasets are either limited to the salient objects of the images, or tend to broadly depict the entire visual scene. A picture is worth a thousand words, and these holistic image descriptions are  far from a complete visual understanding of the image. Furthermore, giving one description for an entire image can sometimes be quite subjective, making the evaluation of captioning often ambiguous. 

Recently, Johnson \etal~\cite{Johnson2015densecap} propose to use a dense description of image regions as a better interpretation of the visual content, known as dense captioning. Human annotators are required to exhaustively label bounding boxes over different levels of visual concepts (e.g., objects, object parts, and interactions between them). Compared to global image descriptions, dense local descriptions are more objective and less affected by annotator preference. The local descriptions provide a rich and dense semantic labeling of the visual elements, which can benefit other tasks such as semantic segmentation~\cite{Long2015fully} and visual question answering~\cite{Antol2015vqa}~\cite{Malinowski2015ask}. For convenience, we refer to image regions associated with annotated visual concepts as \emph{regions of interest} in the remaining text.

The exploration of dense captioning is only just beginning. An end-to-end neural network is used in~\cite{Johnson2015densecap} to predict descriptions based on region proposals generated by a region proposal network~\cite{Ren2015faster}. For each region proposal, the network produces three elements separately: a region-of-interest probability (similar to the detection score in object detection), a phrase to describe the content, and a bounding box offset. The major difference dense captioning has from traditional object detection is that it has an open set of targets (not limited to valid objects), and includes parts of objects and multi-object interactions. Because of this, two types of challenges emerge when predicting region captions.   

\begin{figure}[t]\centering
\includegraphics[width=1\linewidth]{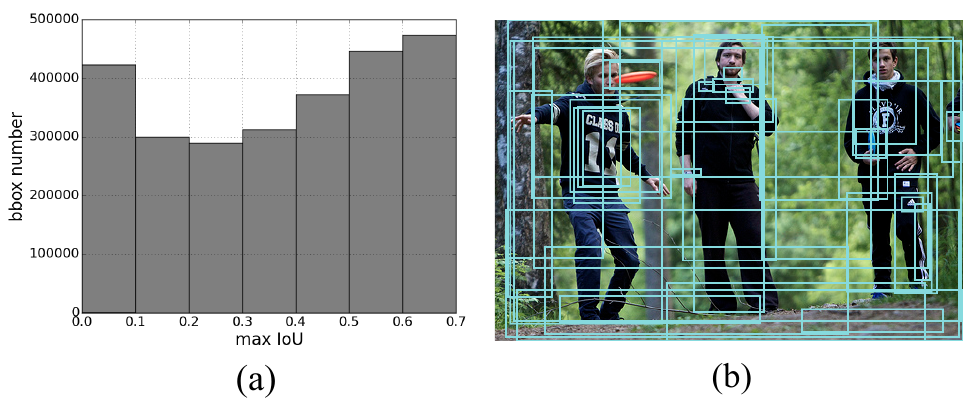}
\caption{(a) Distribution of maximum IoUs between bounding boxes in ground truth annotations. (b) Sample image overlaid with all ground truth bounding boxes. }
\label{fig:dense}
\vspace{-3pt}
\end{figure}

First, the target bounding boxes become much denser than object detection with limited categories (e.g. 20 categories for PASCAL VOC~\cite{pascal-voc-2012}). Take the Visual Genome dataset as an example. The statistics of the maximum Intersection-over-Union (IoU) between ground truth bounding boxes can be seen in Fig.~\ref{fig:dense}(a), from which we see more than half of the bounding boxes have maximum IoU larger than 0.3\footnote{Note that because a large portion of overlapping bounding boxes refer to the same object and have high IoU ratios, we have merged the bounding boxes with IoU larger than 0.7 together into one.}. Fig.~\ref{fig:dense}(b) shows an image overlaid with all ground truth bounding boxes. Here, we can visually see that any region proposal can easily have multiple overlapping regions of interest. Therefore, it is necessary to localize a target region with the guidance of the description. 
%
%

Second, since there are a huge number of visual concepts being described, some of the target regions are visually ambiguous without information about their context. In Visual Genome, the number of different object categories is $18,136$~\cite{Krishna2016visual}, which includes a long list of visually similar object pairs such as ``street light'' and ``pole'',  ``horse'' and ``donkey'', and ``chair'' and ``bench''.  

Thus, we believe that tackling these two challenges can greatly benefit the task of dense captioning. We carefully design our dense captioning model to address the above two problems by introducing two key components. The first component is joint inference, where pooled features from regions of interest are fed into a recurrent neural network to predict region descriptions, and the localization bounding boxes are jointly predicted from the the pooled features with recurrent inputs from the predicted descriptions. Fig.~\ref{fig:overview}(a) shows an example of a step-by-step localization process with joint inference, where the localization bounding box gradually adapts to the correct position using the predicted descriptions. The second component is context fusion, where pooled features from regions of interest are combined with context features to predict better region descriptions. An example is shown in Fig.~\ref{fig:overview}(b), where the object in the red bounding box is described as a desktop without visual cues from the surrounding context. We design several different network structures to implement the two key components respectively, and conduct extensive experiments to explore the benefits and characteristics of each. Our unified model achieves a mean average precision (mAP) accuracy of 9.31\% on Visual Genome V1.0, a relative gain of 73\% over the previous state-of-the-art approach by~\cite{Johnson2015densecap}. An example image with the most confident region descriptions from our model is shown in Fig.~\ref{fig:overview}(c). 

To reiterate, the contributions of this work are two-fold:
\begin{itemize}
\item We design network structures that incorporate two novel ideas, joint inference and context fusion, to address the challenges we identified in dense captioning.

\item We conduct an extensive set of experiments to explore the capabilities of the different model structures, and analyze the underlying mechanisms for each. With this, we are able to obtain a compact and effective model with state-of-the-art performance.
\end{itemize}

\section{Related Work}
%
%
%
%
%
Recent image captioning models often utilize a convolutional neural network (CNN)~\cite{Lecun1998gradient} as an image encoder and a recurrent neural network (RNN)~\cite{Werbos1988generalization} as a decoder for predicting a sentence~\cite{Donahue2015long}~\cite{Karpathy2015deep}~\cite{Vinyals2015show}. RNNs have been widely used in language modeling~\cite{Bengio2003neural}~\cite{Hochreiter1997long}~\cite{Mikolov2010recurrent}~\cite{Sutskever2011generating}. Some image captioning approaches, though targeted at a global description, also build relationships with local visual elements. Karpathy \etal~\cite{Karpathy2015deep}~\cite{Karpathy2014deep} learn an embedding with a latent alignment between image regions and word phrases. Fang \etal~\cite{Fang2015captions} first detect words from images using multiple instance learning, then incorporate the words in a maximum entropy language model. A soft-attention mechanism is also proposed to cast attention over different image regions when predicting each word~\cite{Jin2015aligning}~\cite{Xu2015show}. 

Recent object detection algorithms based on deep learning often show a two-stage paradigm: region proposal and detection~\cite{Girshick2015fast}~\cite{Girshick2014rich}~\cite{Ren2015faster}. Faster R-CNN~\cite{Ren2015faster} is the most related to our work, as it utilizes a Region Proposal Network (RPN) to generate region proposals and a detection network to predict object categories and bounding box offsets. The two networks can share convolutional features and can be trained with an approximate fast joint training algorithm. A recent improvement to faster R-CNN is the incorporation of context information using a four-direction RNN on the convolutional feature map~\cite{Bell2015inside}. Visual context can greatly help tasks such as object detection~\cite{Bell2015inside}~\cite{Divvala2009empirical}~\cite{Mottaghi2014role} and semantic segmentation~\cite{Mottaghi2014role}. Another direction is to remove the RPN and directly produce detection results~\cite{Liu2015ssd}~\cite{Redmon2015you} to further speed up the algorithm.

The task of dense captioning was first proposed in~\cite{Johnson2015densecap}, in which a spatial transformer network~\cite{Jaderberg2015spatial} is used to facilitate joint training of the whole network. A related application is also proposed to detect an arbitrary phrase in images using the dense captioning model. The experiments are conducted on the Visual Genome dataset~\cite{Krishna2016visual}, which provides not only region descriptions but also objects, attributes, question answering pairs, etc. 
Also closely related are other recent topics such as localizing a phrase in a specific image~\cite{Hu2015natural}~\cite{Mao2016unambiguous}~\cite{Nagaraja2016refer}, generating an unambiguous description for a specific region in an image~\cite{Mao2016unambiguous}~\cite{Yu2016refer}, or detecting visual relationships in images~\cite{Li2017ViPCNN}~\cite{Lu2016visual}. 

\section{Our Model}

\begin{figure}[t]\centering
\includegraphics[width=1\linewidth]{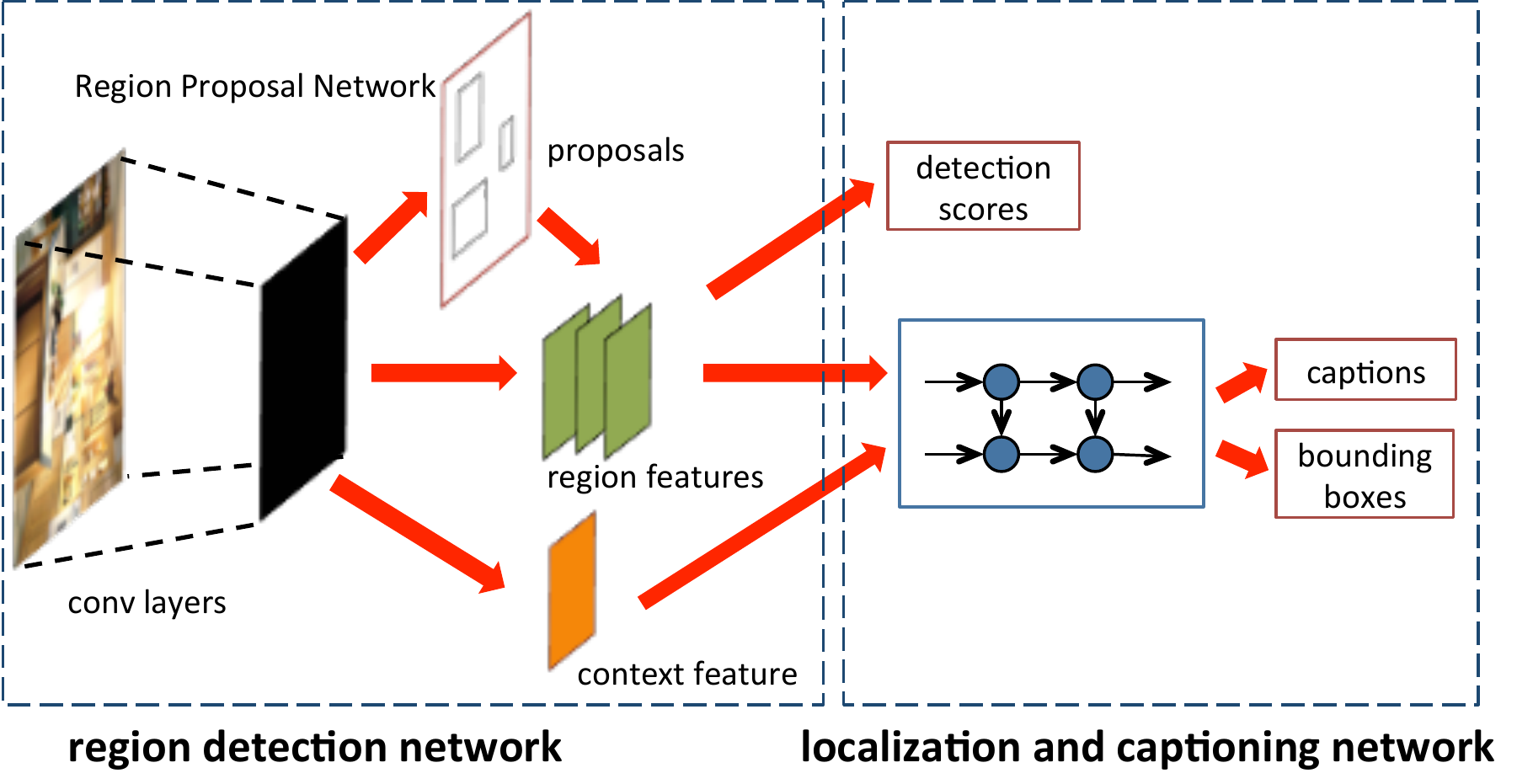}
\caption{Our framework consists of two stages: a region detection network and a localization and captioning network.}
\label{fig:framework}
\vspace{-3pt}
\end{figure}

Dense captioning is similar to object detection in that it also needs to localize the regions of interest in an image, but differs in that it replaces the fixed number of object categories with a much larger set of visual concepts described by phrases. Therefore, we can borrow successful recipes from the object detection literature in designing our dense captioning algorithm. In this work, our dense captioning model pipeline is inspired by the efficient faster R-CNN framework~\cite{Ren2015faster}. Fig.~\ref{fig:framework} illustrates our dense captioning framework, which includes a region detection network adopted from faster R-CNN and a localization and captioning network. In this section, we will design different localization and captioning network architectures step by step in searching for the right formula. Our baseline model directly combines the faster R-CNN framework for region detection and long short-term memory (LSTM)~\cite{Hochreiter1997long} for captioning. 

\begin{figure}[t]\centering
\includegraphics[width=0.8\linewidth]{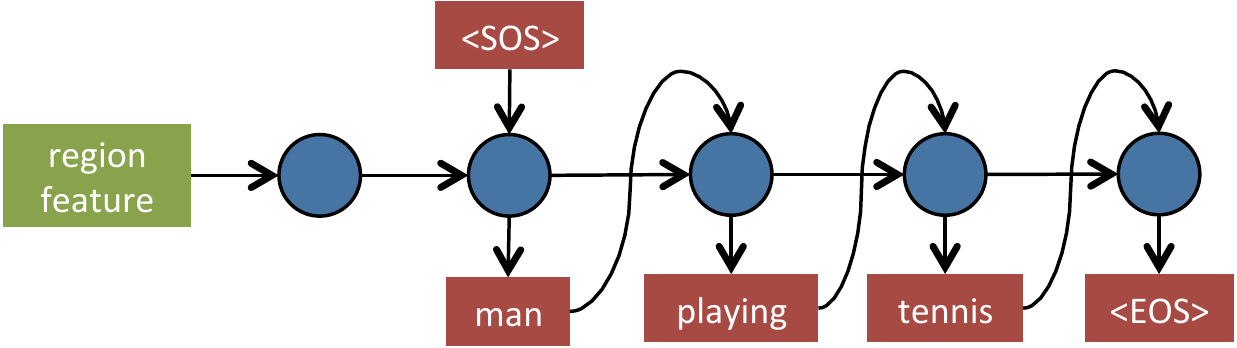}
\caption{An illustration of the unrolled LSTM for region captioning. $<$SOS$>$ and $<$EOS$>$ denote the start-of-sentence and end-of-sentence tokens, respectively.}
\label{fig:lstm}
\vspace{-3pt}
\end{figure}

\begin{figure*}[t]\centering
\includegraphics[width=0.8\linewidth]{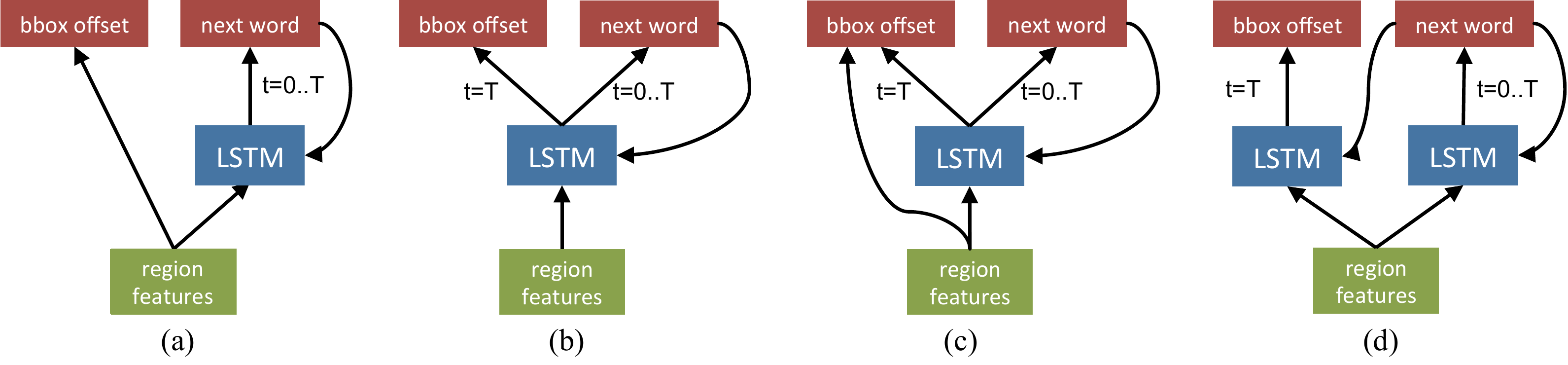}
\caption{Baseline model and several model designs for joint inference of bounding box offset and region description. The four structures are (a) Baseline model (b) S-LSTM (c) SC-LSTM (d) T-LSTM. See detailed description in text.}
\label{fig:joint}
\vspace{-3pt}
\end{figure*}

\subsection{Baseline model}

Faster R-CNN~\cite{Ren2015faster} uses a two-stage neural network to detect objects based on the image feature maps, which are generated by a fully convolutional neural network. In the first stage, the network uses a RPN to generate region proposals that are highly likely to be the regions of interest, then it generates fixed-length feature vectors for each region proposal using Region-Of-Interest (ROI) pooling layers. In the second stage, the feature vectors are fed into another network to predict object categories as well as the bounding box offsets. Since the gradients cannot be propagated through the proposal coordinates, exact joint training is not viable for faster R-CNN. Instead, it can be trained by alternatively updating parameters with gradients from the RPN and the final prediction network, or by approximate joint training which updates the parameters with gradients from the two parts jointly.

Our baseline model for dense captioning directly uses the proposal detection network from faster R-CNN in the first stage. For the second stage of localization and captioning, we use the model structure in Fig.~\ref{fig:joint}(a), where the region features are used to produce detection scores and bounding box offsets, as well as fed into an LSTM to generate region descriptions. The LSTM predicts a word at each time step and uses this prediction to predict the next word. Fig.~\ref{fig:lstm} shows an example of using such a recurrent process to generate descriptions.
We use the structure of VGG-16~\cite{Simonyan2014very} for the convolutional layers, which generates feature maps $16\times$  smaller than the input image. Following faster R-CNN~\cite{Ren2015faster}, pretrained weights from the ImageNet Classification challenge~\cite{Deng2009imagenet} are used. Also following previous work~\cite{Johnson2015densecap} ~\cite{Karpathy2015deep}~\cite{Mao2014explain}~\cite{Vinyals2015show}, the region feature is only fed into the LSTM at the first time step, followed by a special start-of-sentence token, and then by the embedded feature vectors of the predicted words one by one. This model is similar to the model in~\cite{Johnson2015densecap} except that their model replaces the ROI pooling layer with a bilinear interpolation module so that gradients can be propagated through bounding box coordinates. In contrast, our baseline model uses approximate joint training that is proven to be effective for object detection and instance-level semantic segmentation~\cite{Jiang2016face}~\cite{Dai2016instance}~\cite{Ren2015faster}. In our experiments, we observe that the baseline model with approximate joint training is very effective and already outperforms the previous state-of-the-art method~\cite{Johnson2015densecap}. A potential reason is that although bilinear interpolation allows for exact end-to-end training, the model may be harder to train due to the transformation properties of the gradients.

\subsection{Joint inference for accurate localization}\label{sec:joint}
In this section, we  explore the model design for joint inference of bounding box localization and region description for a given region proposal. Due to the large number of open phrases and dense bounding boxes, we find it is necessary to combine the two in order to improve both localization and captioning. We fix the first stage of the proposal detection network in Fig.~\ref{fig:framework} to be the same as our baseline model, and focus on designing a joint localization and captioning network for the second stage.

To make the predictor of the bounding box offset aware of the semantic information in the associated region, we make the bounding box offset an output of an LSTM encoded with region descriptions. Several designs are shown in Fig.~\ref{fig:joint}. Shared-LSTM (S-LSTM) (Fig.~\ref{fig:joint}(b)) directly uses the existing LSTM to predict the bounding box offset at the last time step of the caption. This model embeds the captioning model and the location information in the same hidden space. Shared-Concatenation-LSTM (SC-LSTM) (Fig.~\ref{fig:joint}(c)) concatenates the output of the LSTM and region features to predict the offset, so the prediction of the offset is directly guided by the region features. Twin-LSTM (T-LSTM) (Fig.~\ref{fig:joint}(d)) uses two LSTMs to predict the bounding box offset and description separately. This model separates the embedded hidden spaces of the captioning model and the location predictor. The two LSTMs are denoted as \emph{location-LSTM} and \emph{caption-LSTM}, and both receive the embedded representation of the last predicted word as input. In all three models, the bounding box offset is predicted at the last time step of the description, when the ``next word'' is an end-of-sentence token and the description is finished. Thus the network obtains information about the whole description at the time of predicting the bounding box offset.

\subsection{Context fusion for accurate description}\label{sec:context}

Visual context is important for understanding a local region in an image, where it has already shown to benefit tasks such as object detection and semantic segmentation~\cite{Bell2015inside}~\cite{Divvala2009empirical}~\cite{Mottaghi2014role}. Despite the exploration of context features in these tasks, there is limited work on the integration of context features into sequential prediction tasks such as image captioning. We concentrate on finding the optimal way to combine context features and local features in the sequential prediction task of dense captioning, rather than investigating better representations of context information. Thus, we resort to a simple but effective implementation of context features, which utilizes a global ROI pooling feature vector as the visual context. Since the bounding box offset is not directly connected to the context feature, we only use the context feature to assist in caption prediction, which in turn will influence localization through joint inference as discussed in the previous section. 

\begin{figure}[t]\centering
\includegraphics[width=0.72\linewidth]{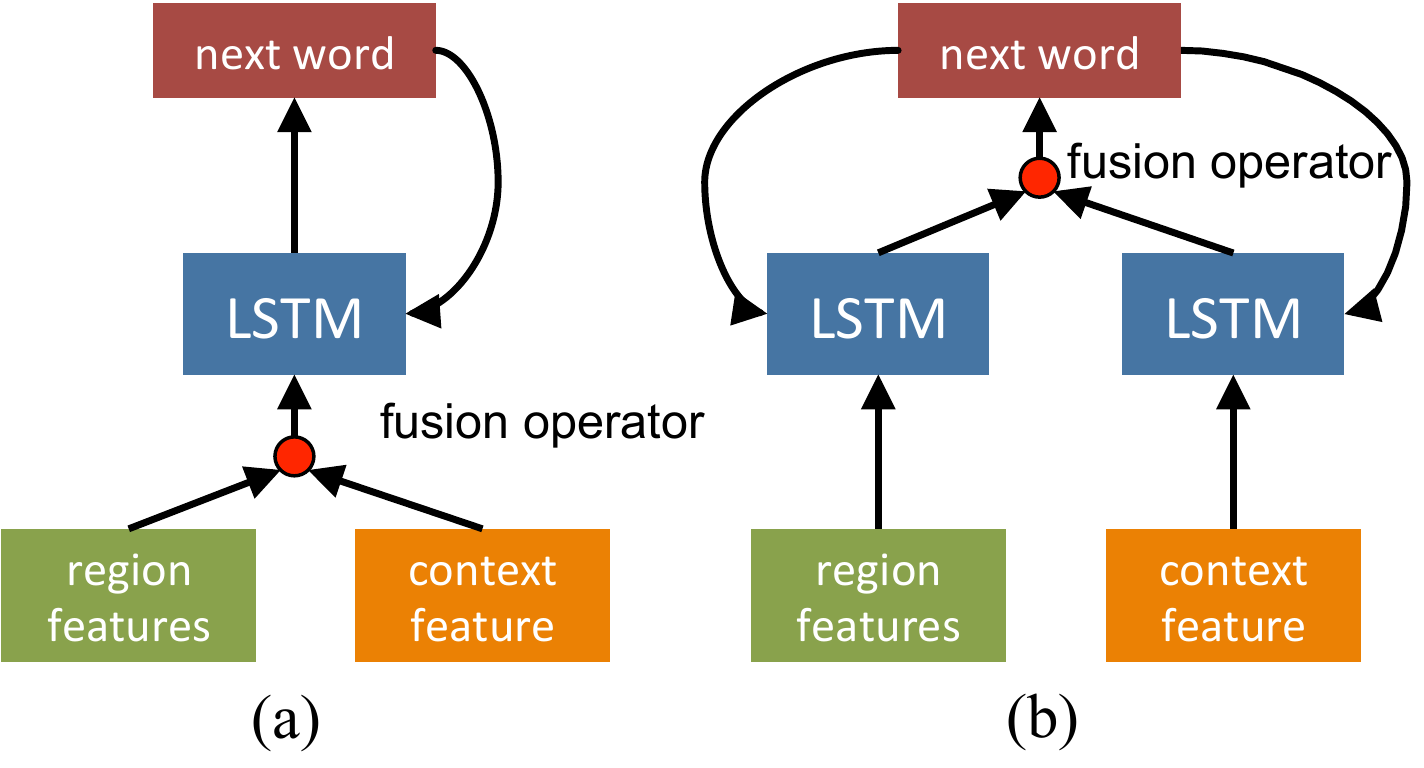}
\caption{Model structures for region description assisted by context features. (a) Early-fusion. (b) Late-fusion. The fusion operator denoted by the red dot can be concatenation, summation, multiplication, etc.}
\label{fig:context}
\vspace{-3pt}
\end{figure}

In this work, we experiment with two variants of combining local features and context features, which are shown in Fig.~\ref{fig:context} and termed as \emph{early-fusion} and \emph{late-fusion}. Early-fusion (Fig.~\ref{fig:context}(a)) directly combines the region feature and context feature together before feeding into the LSTM, while late-fusion (Fig.\ref{fig:context}(b)) uses an extra LSTM to generate a recurrent representation of the context feature, and then combines it with the local feature. The context feature representation is combined with the region feature representation via a \emph{fusion operator} for both variants. We experimented with concatenation, summation, and multiplication. After each word is selected, its embedded representation is fed back into the caption-LSTM to guide the generation of the next word. Such fusion designs can be easily integrated with any of the models in Fig.~\ref{fig:joint}.

\subsection{Integrated model}
\begin{figure}[t]\centering
\begin{subfigure}[b]{0.25\textwidth}
   \includegraphics[width=1\linewidth]{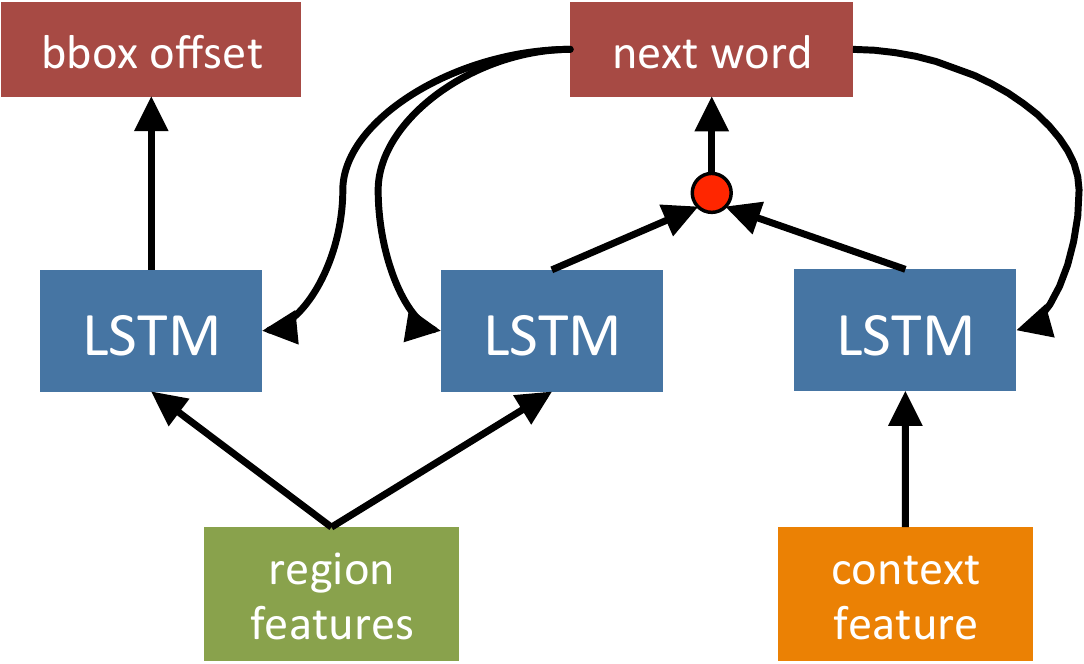}
   \subcaption{}
\end{subfigure}
\begin{subfigure}[b]{0.35\textwidth}
\includegraphics[width=1\linewidth]{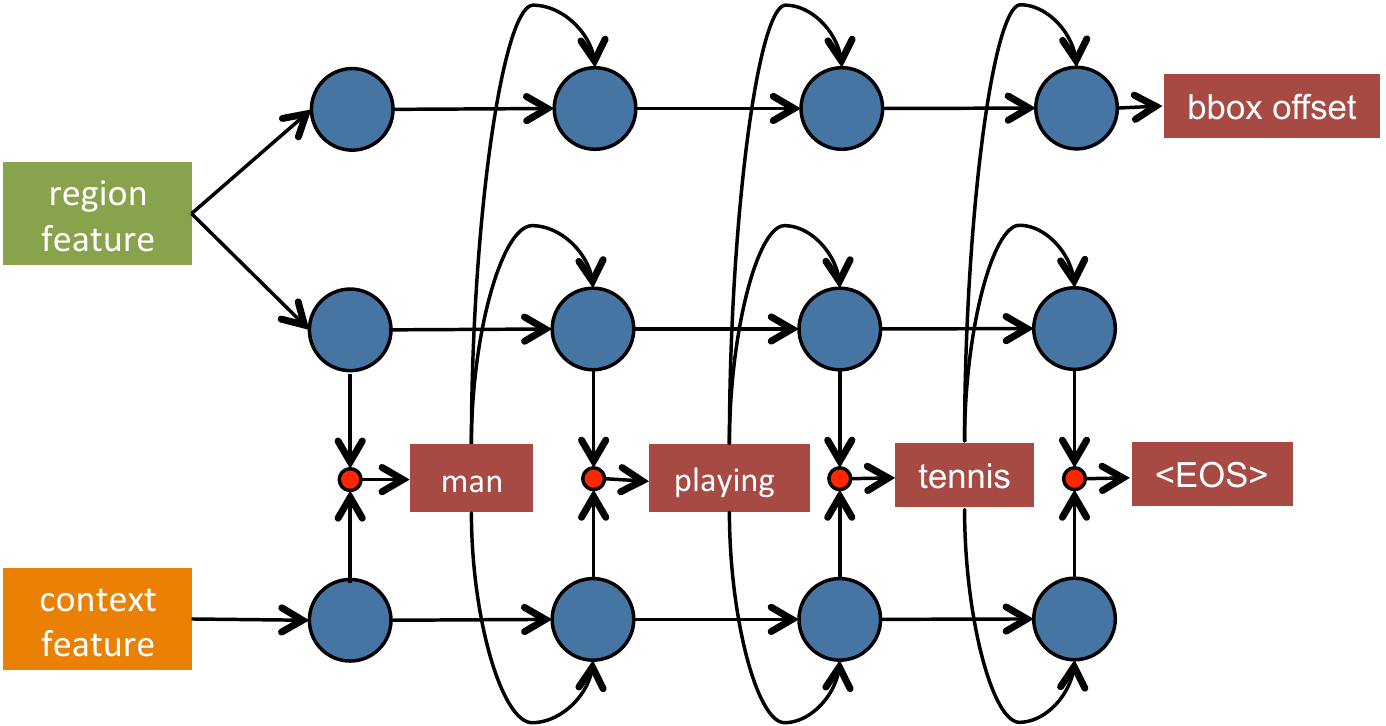}
\subcaption{}
\vspace{-3pt}
\end{subfigure}
\caption{Illustrations of an integrated model. (a) The integrated model of T-LSTM and the late-fusion context model. (b) An unrolled example of the model in (a). Start-of-sentence token is omitted for clarity.}
\label{fig:two-lstm-context-late}
\vspace{-3pt}
\end{figure}

The aforementioned model structures of joint inference and context fusion can be easily plugged together to produce an integrated model. For example, the integration of T-LSTM and the late-fusion context model can be viewed in Fig.~\ref{fig:two-lstm-context-late}. Note that a single word is predicted at each time step and the bounding box offset is predicted at the last time step of the caption, after all words have been encoded into the location-LSTM. Different integrated models are different instantiations of the model pipeline we show in Fig.~\ref{fig:framework}. 

Finally, training our dense captioning model boils down to minimizing the following loss function $L$,
\begin{align}
L = L_{cap} + \alpha L_{det} + \beta L_{bbox},
\label{eq:loss}
\end{align}
where$L_{cap}$, $L_{det}$, and $L_{bbox}$ denote caption prediction loss, detection loss and bounding box regression loss, respectively, with $\alpha$ and $\beta$ the weighting coefficients.  $L_{cap}$ is a cross-entropy term for word prediction at each time step of the sequential model, $L_{det}$ is a two-class cross-entropy loss for foreground / background regions, while $L_{bbox}$ is a smoothed-L1 loss~\cite{Ren2015faster}.  $L_{det}$ and $L_{bbox}$ are computed both in the region proposal network and the final prediction. For those models using an LSTM for predicting bounding box offset, the second $L_{bbox}$ is calculated at the last time-step of the LSTM output. 

\section{Experiments}

\subsection{Evaluation Dataset}

We use the Visual Genome dataset~\cite{Krishna2016visual} as the evaluation benchmark. Visual Genome has two versions: V1.0 and V1.2. V1.2 is a cleaner version of V1.0, while V1.0 is used by~\cite{Johnson2015densecap}. For comparison purposes, we conduct experiments mainly on V1.0, and report additional results on V1.2. We use the same train/val/test splits as in~\cite{Johnson2015densecap} for both V1.0 and V1.2, i.e., 77398 images for training and 5000 images each for validation and test. We use the same evaluation metric of mean Average Precision (mAP) as~\cite{Johnson2015densecap}, which measures localization and description accuracy jointly. Average precision is computed for different IoU thresholds for localization accuracy, and different Meteor~\cite{Banerjee2005meteor} score thresholds for language similarity, then averaged to produce the mAP score. For localization, IoU thresholds .3, .4, .5, .6, .7 are used. For language similarity, Meteor score thresholds 0, .05, .1, .15, .2, .25 are used.
A comparison of our final model using the structure in Fig.~\ref{fig:two-lstm-context-late} with the previous best result can be seen in Tab.~\ref{tab:best}, which shows that we achieve a $73\%$ relative gain compared to the previous best method.
In the following sections, we first introduce the training and evaluation details, then evaluate and compare the joint inference models and integrated models under different structure designs.
The influence of hyper-parameters in evaluation is also explored.

\begin{table}[]
\centering
\footnotesize
\caption{Comparison of our final model with previous best result on Visual Genome V1.0 and V1.2.}
\label{tab:best}
\begin{tabular}{|l|l|l|l|l|}
\hline
& \multicolumn{3}{|l|}{Visual Genome V1.0} & V1.2 \\ \hline 
 Model   & Johnson \etal~\cite{Johnson2015densecap} & Ours& Gain  & Ours \\ \hline
mAP & 5.39    & \bf{9.31}  & 73\%            &  \bf{9.96}    \\ \hline
\end{tabular}
\vspace{-3pt}
\end{table}

\begin{table*}[]
\footnotesize
\centering
\caption{The mAP performance of baseline and joint inference models on Visual Genome V1.0. First row is the performance with CNN and RPN fixed, second row is the performance of corresponding models with end-to-end training.}
\label{tab:results_joint}
\begin{tabular}{|l|l|l|l|l|l|}
\hline
 model     & Johnson \etal~\cite{Johnson2015densecap} & baseline              & S-LSTM & SC-LSTM & T-LSTM \\ \hline
 fixed-CNN\&RPN   & - & 5.26 & 5.15     & 5.57           & 5.64           \\ \hline
end-to-end    & 5.39  & 6.85 & 6.47     & 6.83           & \textbf{8.03}           \\ \hline      
\end{tabular}
\vspace{-3pt}
\end{table*}

\begin{table}[]
\footnotesize
\centering
\caption{The mAP performance of integrated models with combinations of joint inference models and context fusion structures on Visual Genome V1.0.}
\label{tab:results_integrated}
\begin{tabular}{|l|l|l|l|l|}
\hline
\multicolumn{2}{|l|}{model}             & S-LSTM & SC-LSTM & T-LSTM \\ \hline
\multirow{2}{*}{early-fusion} & $[\cdot,\cdot]$&  6.74 &  7.18 & 8.24 \\ \cline{2-5}
                            & $\oplus$  &  6.54        &  7.29              &  8.16        \\ \cline{2-5} 
                            & $\otimes$ &  6.69        &  7.04              & 8.19         \\ \hline
\multirow{3}{*}{late-fusion}  & $[\cdot,\cdot]$ & 7.50         &   7.72             &   8.49       \\ \cline{2-5} 
                              & $\oplus$  &    7.19      &   7.47             &  8.53        \\ \cline{2-5} 
                              & $\otimes$ &   7.57       &   7.64             &  \textbf{8.60}        \\ \hline
\end{tabular}
\vspace{-3pt}
\end{table}

\begin{table}[]
\footnotesize
\centering
\caption{The mAP performance of different dense captioning models on Visual Genome V1.2.}
\label{tab:results_1.2}
\begin{tabular}{|l|l|l|l|l|}
\hline
\multicolumn{2}{|l|}{model}                     & baseline              & S-LSTM & T-LSTM \\ \hline
\multicolumn{2}{|l|}{no context}              & \multirow{4}{*}{6.98} & 6.44     & 8.16     \\ \cline{1-2} \cline{4-5} 
\multirow{3}{*}{late-fusion} & $[\cdot,\cdot]$ &                    &  7.76          &  \textbf{9.03}        \\ \cline{2-2} \cline{4-5} 
                              & $\oplus$     &                    &  7.06           &   8.71       \\ \cline{2-2} \cline{4-5} 
                              & $\otimes$       &                       &  7.63        & 8.52         \\ \hline
\end{tabular}
\vspace{-3pt}
\end{table}

\subsection{Model training and evaluation}

In training, we use approximate joint training for all models. We use stochastic gradient descent with a mini-batch size of 1 to train the whole network. The input image is re-sized to have a longer side of $720$ pixels. Initial learning rate is set to $0.001$ and halved every $100$K iterations, and momentum is set to $0.98$. Weight decay is not used in training. We begin fine-tuning the CNN layers after $200$K iterations ($\sim$3 epochs) and finish training after $600$K iterations ($\sim$9 epochs). The first seven convolutional layers are fixed for efficiency, with the other convolutional layers fine-tuned. We found that training models with context fusion from scratch tends not to converge well, so we fine-tune these models from their non-context counterparts, with a total of $600$K training iterations. We only use descriptions with no more than 10 words for efficiency. We use the most frequent $10000$ words as the vocabulary and replace other words with an $<$UNK$>$ tag. For sequential modeling, We use an LSTM with $512$ hidden nodes. For the RPN, we use $12$ anchor boxes for generating the anchor positions in each cell of the feature map, and $256$ boxes are sampled in each forward pass of training. For the loss function, we fix values of $\alpha$, $\beta$ in Eq.~(\ref{eq:loss}) to $0.1$ and $0.01$, respectively. 

In evaluation, we follow the settings of~\cite{Johnson2015densecap} for fair comparison. First, $300$ boxes with the highest predicted confidence after non-maximum suppression (NMS) with IoU ratio $0.7$ are generated. Then, the corresponding region features are fed into the second stage of the network, which produces detection scores, bounding boxes, and region descriptions. We use efficient beam-1 search to produce region descriptions, where the word with the highest probability is selected at each time step. With another round of NMS with IoU ratio $0.3$, the remaining regions and their descriptions are used as the final results.

\subsection{Joint inference models}
We evaluate the baseline and three joint inference models in this section. All models are trained end-to-end with the convolutional layers and the RPN. 
To further clarify the effect of different model designs, we also conduct experiments to evaluate the performance of the models based on the same region proposals and image features. Towards this end, we fix the weights of the CNN to those of VGG16 and use a hold-out region proposal network also trained on Visual Genome based on the fixed CNN weights. The results of the end-to-end trained models and the fixed-CNN\&RPN models are shown in Tab.~\ref{tab:results_joint}.

%
%
%
\begin{figure*}[t]\centering
\includegraphics[width=1\linewidth]{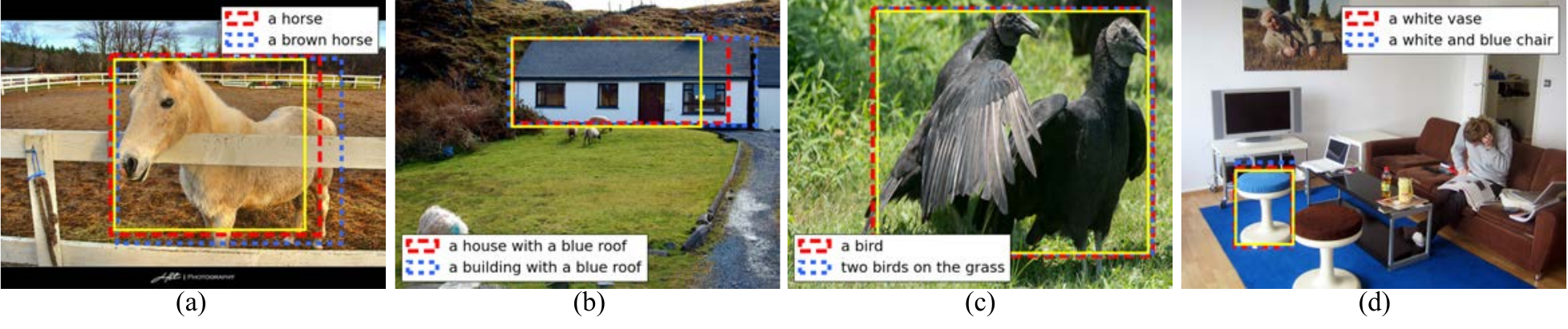}
\caption{Qualitative comparisons between baseline and T-LSTM. In each image, the yellow box, the red box, and the blue box are the region proposal, the prediction of the baseline model, and the prediction of the T-LSTM model, respectively. }
\label{fig:compare1}
\vspace{-5pt}
\end{figure*}

\textbf{T-LSTM performs best for joint inference.} Among the three different structures for joint inference, T-LSTM has the best performance for both end-to-end training (mAP 8.03), and fixed-CNN\&RPN training (mAP 5.64). The end-to-end model of T-LSTM outperforms the baseline model by more than 1\% in mAP, while the others are even worse than the baseline model. By using a shared LSTM to predict both the caption and bounding box offset, S-LSTM unifies the language representation and the target location information into a single hidden space, which is quite challenging since they are from completely different domains. Even assisted by the original region feature, the shared LSTM solution does not show much improvement, only on par with the baseline (mAP 6.83). By separating the hidden space, i.e. using two LSTMs targeted at the two tasks respectively, the T-LSTM model yields much better performance (mAP 8.03 vs 6.47).  Compared with the baseline model, T-LSTM is better at both localization and captioning.  Fig.~\ref{fig:compare1} shows several example predictions of bounding box and captions from one region proposal for the baseline model and the T-LSTM model. Fig.~\ref{fig:compare1}(a)~(b) shows that T-LSTM improves on localization thanks to the guidance of the encoded caption information, while Fig.~\ref{fig:compare1}(c)~(d) shows that T-LSTM is also better at predicting the descriptions, which reveals that location information helps to improve captioning. Although bounding box prediction does not feed information to the captioning process in the forward pass, it does influence captioning through back-propagation in the training stage.
Considering all these factors, we believe that separating the hidden space using T-LSTM is most suitable for the joint inference of caption and location.


\subsection{Integrated models}
We evaluate the integrated models with different designs for both joint inference and context fusion in this section. For joint inference models, we evaluate three variants: S-LSTM, SC-LSTM, and T-LSTM. For context fusion, we compare the different settings proposed in Section~\ref{sec:context}, where we evaluate early-fusion and late-fusion with different fusion operators: concatenation, summation, and multiplication. For early-fusion with concatenation, we plug in a fully-connected layer after the concatenated feature to reduce it to the same input dimension as the LSTM. The mAP results of different variants of models are shown in Tab.~\ref{tab:results_integrated}.

\begin{figure*}[t]\centering
\includegraphics[width=1\linewidth]{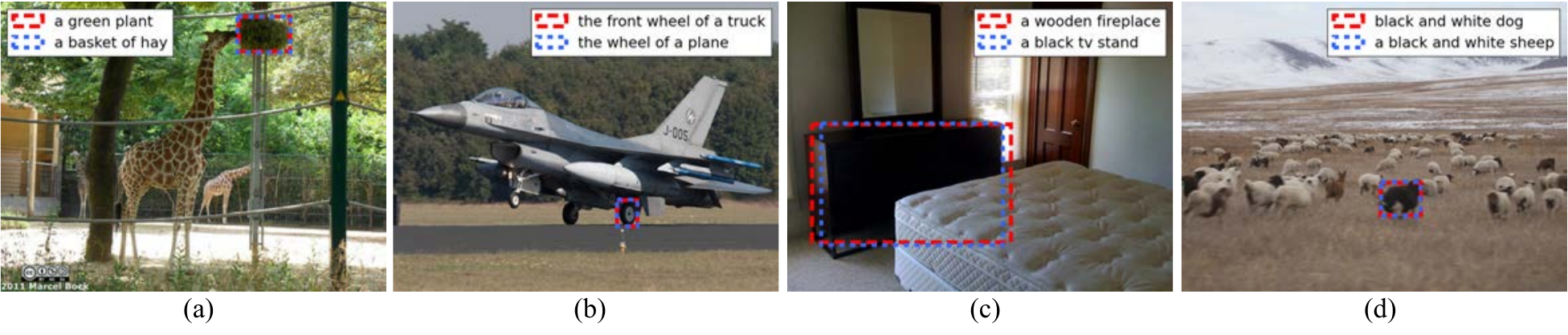}
\caption{Qualitative comparisons of T-LSTM and T-LSTM-mult. In each image, the red box and the blue box are the prediction of the no-context and context model, respectively, with their predicted captions. Region proposals are omitted.}
\label{fig:compare2}
\vspace{-3pt}
\end{figure*}

\textbf{Effectiveness of context fusion.}
In all models, context information helps to improve mAP ranging from 0.07 (S-LSTM, early-fusion, summation) to 1.10 (S-LSTM, late-fusion, multiplication). The three types of fusion methods all yield improvements in mAP for different models. Generally, concatenation and multiplication are more effective than summation, but the margin is subtle. With T-LSTM and late-fusion with multiplication, we obtain the best mAP performance 8.60 in this set of experiments. We refer to this model as T-LSTM-mult for brevity in the remaining text.
Fig.~\ref{fig:compare2} shows example predictions for comparison of T-LSTM without context fusion and T-LSTM-mult. In Fig.~\ref{fig:compare2}(a)~(b)~(c), T-LSTM-mult gives a better caption than the model without context. Without context, these objects are very hard to recognize even by humans. We can also observe from these examples that the context information employed by the model is not limited to the surrounding part of the region proposal, but from the whole image. In Fig.~\ref{fig:compare2}(d), the context model interestingly gives an incorrect but reasonable prediction since it is misled by the context which is full of sheep. 

\textbf{Late-fusion is better than early-fusion.}
Comparing early-fusion and late-fusion of context information, we find that late-fusion is better than early-fusion for all pairs of corresponding models. Also, early fusion only outperforms its no-context counterparts by a small margin. One disadvantage of early-fusion is that it directly combines the local and context features that have quite differing visual elements, making it unlikely able to decorrelate the visual element into the local region or the context region in the later stages of the model.

\begin{figure*}[t]\centering
\includegraphics[width=1\linewidth]{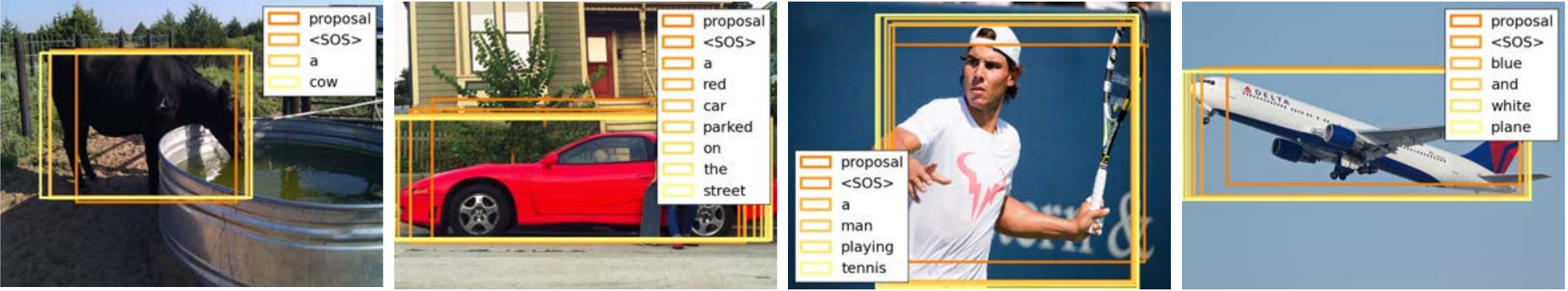}
\caption{Bounding box predictions at different time steps of the caption using T-LSTM-mult. In each image, different colors of boxes denote the outputs of different time steps, with the brighter the color the later in time. The corresponding words fed into the location-LSTM are shown in the legends. $<$SOS$>$ is the start-of-sentence token.}
\label{fig:timesteps}
\vspace{-5pt}
\end{figure*}

\textbf{Intermediate location predictions.} Since we only add the regression target to the last time step of the location-LSTM, it is not clear what the bounding box predictions from the previous time steps will be like. We test the bounding box predictions from these time steps, and find them to be fairly good. Fig.~\ref{fig:timesteps} shows examples of the predicted bounding box location at different time steps for the T-LSTM-mult model. Generally, the bounding box prediction at the first time step is already close to the region of interest. As words are fed into the location-LSTM, it gradually adjusts the bounding box to a tight localization of the object being described. 
%
%
%

\begin{table}[]
\small
\centering
\caption{The chosen hyper-parameters and the performance on Visual Genome V1.0 and V1.2 respectively.}
\label{tab:cv}
\begin{tabular}{|l|l|l|l|l|}
\hline
&\#proposal & NMS\_r1 & NMS\_r2 & mAP \\ \hline
\multirow{2}{*}{V1.0} &100     &   0.5    &   0.4     & 8.67    \\ \cline{2-5}
&300        &  0.6      & 0.5        & \textbf{9.31}     \\ \hline
\multirow{2}{*}{V1.2} &100     &  0.5      &    0.5    & 9.47    \\ \cline{2-5}
&300        &  0.6      & 0.5       & \textbf{9.96}    \\ \hline
\end{tabular}
\vspace{-3pt}
\end{table}

\subsection{Results on Visual Genome V1.2}
We also conduct experiments on Visual Genome V1.2 using the same train/val/test split as V1.0. The mAP performances are shown in Tab.~\ref{tab:results_1.2}. Here, we see similar results as on V1.0, which further verifies the advantage of T-LSTM over S-LSTM (mAP 8.16 vs 6.44 for no-context), and that context fusion greatly improves performance for both models. For context fusion, we can see that the T-LSTM model with late concatenation achieves the best result with mAP $9.03$. We refer to this model as T-LSTM-concat. Overall, the accuracies are higher than those on Visual Genome V1.0, likely due to the cleaner ground truth labels. 

\subsection{Best practice: hyper-parameters}
The evaluation pipeline for dense captioning, a two-stage process of target prediction (region proposal and region description along with location refinement), involves several hyper-parameters that can influence the accuracy. These parameters include the number of proposals given by the RPN and the IoU ratio used by NMS both in the RPN and the final prediction. For these parameters, we use the same settings as~\cite{Johnson2015densecap} for all evaluations above. However, we are also interested in the impact of these parameters on our results. Parameters such as number of proposals is worth investigating because it can be used to find a trade-off between speed and performance. Also, the NMS thresholds used by~\cite{Johnson2015densecap} seem to overly suppress the predicted bounding box, especially since the ground truth regions are very dense (Fig.~\ref{fig:dense}). 

We use T-LSTM-mult for Visual Genome V1.0 and T-LSTM-concat for V1.2 as prototypes and find the best hyper-parameters for each by validating on the validation set. For the number of proposals given by the RPN, we validate between 100 and 300 proposals. We also validate to find the optimal IoU ratios used in the NMS thresholds for RPN and for final prediction, denoted as NMS\_r1 and NMS\_r2, respectively. NMS\_r1 is chosen from the range $0.4 \sim 0.9$, and NMS\_r2 is chosen from the range $0.3 \sim 0.8$. The results and corresponding optimal hyper-parameter settings are shown in Tab.~\ref{tab:cv}. 

With the validated hyper-parameters, we achieve even better mAP performance with $9.31$ on Visual Genome V1.0 and $9.96$ on Visual Genome V1.2 using $300$ proposals, which sets the new state-of-the-art. With only $100$ proposals, our model achieves mAP $8.67$ on Visual Genome V1.0 and $9.47$ on Visual Genome V1.2. Comparing the running times, we find that a $600\times720$ image takes $350$ms and $200$ms for $300$ and $100$ proposals on a GTX TITAN GPU, respectively. The LSTM computations take around $30\%$ of the total time consumption. Our implementation is developed using Caffe~\cite{Jia2014caffe}.

\section{Conclusions}
In this work, we have proposed a novel model structure which incorporates two ideas, joint inference and context fusion, to address specific challenges in dense captioning. To find an exact model realization incorporating these two approaches, we design our model step by step and propose different variants for each component. We evaluate the different models extensively, and gain intuitions on the effectiveness of each component and its variants. Finally, we find a model which utilizes the two approaches effectively and achieves state-of-the-art performance on the Visual Genome dataset. The feature representation learned by these models can potentially benefit other computer vision tasks requiring dense visual understanding such as object detection, semantic segmentation, and caption localization. The extensive comparison of different model structures we conducted can hopefully help guide model design in other tasks involving sequential modeling. 

{\small
\bibliographystyle{ieee}
\bibliography{egbib}

\begin{thebibliography}{10}\itemsep=-1pt

\bibitem{Antol2015vqa}
S.~Antol, A.~Agrawal, J.~Lu, M.~Mitchell, D.~Batra, C.~Lawrence~Zitnick, and
  D.~Parikh.
\newblock Vqa: Visual question answering.
\newblock In {\em ICCV}, 2015.

\bibitem{Banerjee2005meteor}
S.~Banerjee and A.~Lavie.
\newblock Meteor: An automatic metric for mt evaluation with improved
  correlation with human judgments.
\newblock In {\em ACL Workshop}, 2005.

\bibitem{Bell2015inside}
S.~Bell, C.~L. Zitnick, K.~Bala, and R.~Girshick.
\newblock Inside-outside net: Detecting objects in context with skip pooling
  and recurrent neural networks.
\newblock {\em CVPR}, 2016.

\bibitem{Bengio2003neural}
Y.~Bengio, R.~Ducharme, P.~Vincent, and C.~Jauvin.
\newblock A neural probabilistic language model.
\newblock {\em JMLR}, 3(Feb):1137--1155, 2003.

\bibitem{Chen2015coco}
X.~Chen, H.~Fang, T.-Y. Lin, R.~Vedantam, S.~Gupta, P.~Doll{\'a}r, and C.~L.
  Zitnick.
\newblock Microsoft coco captions: Data collection and evaluation server.
\newblock {\em arXiv preprint arXiv:1504.00325}, 2015.

\bibitem{Deng2009imagenet}
J.~Deng, W.~Dong, R.~Socher, L.-J. Li, K.~Li, and L.~Fei-Fei.
\newblock Imagenet: A large-scale hierarchical image database.
\newblock In {\em CVPR}, 2009.

\bibitem{Divvala2009empirical}
S.~K. Divvala, D.~Hoiem, J.~H. Hays, A.~A. Efros, and M.~Hebert.
\newblock An empirical study of context in object detection.
\newblock In {\em CVPR}, 2009.

\bibitem{Donahue2015long}
J.~Donahue, L.~Anne~Hendricks, S.~Guadarrama, M.~Rohrbach, S.~Venugopalan,
  K.~Saenko, and T.~Darrell.
\newblock Long-term recurrent convolutional networks for visual recognition and
  description.
\newblock In {\em CVPR}, 2015.

\bibitem{pascal-voc-2012}
M.~Everingham, L.~Van~Gool, C.~K.~I. Williams, J.~Winn, and A.~Zisserman.
\newblock The {PASCAL} {V}isual {O}bject {C}lasses {C}hallenge 2012 {(VOC2012)}
  {R}esults.
\newblock
  http://www.pascal-network.org/challenges/VOC/voc2012/workshop/index.html.

\bibitem{Fang2015captions}
H.~Fang, S.~Gupta, F.~Iandola, R.~K. Srivastava, L.~Deng, P.~Doll{\'a}r,
  J.~Gao, X.~He, M.~Mitchell, J.~C. Platt, et~al.
\newblock From captions to visual concepts and back.
\newblock In {\em CVPR}, 2015.

\bibitem{Girshick2015fast}
R.~Girshick.
\newblock Fast r-cnn.
\newblock In {\em ICCV}, 2015.

\bibitem{Girshick2014rich}
R.~Girshick, J.~Donahue, T.~Darrell, and J.~Malik.
\newblock Rich feature hierarchies for accurate object detection and semantic
  segmentation.
\newblock In {\em CVPR}, 2014.

\bibitem{Hochreiter1997long}
S.~Hochreiter and J.~Schmidhuber.
\newblock Long short-term memory.
\newblock {\em Neural computation}, 9(8):1735--1780, 1997.

\bibitem{Hu2015natural}
R.~Hu, H.~Xu, M.~Rohrbach, J.~Feng, K.~Saenko, and T.~Darrell.
\newblock Natural language object retrieval.
\newblock {\em CVPR}, 2016.

\bibitem{Jaderberg2015spatial}
M.~Jaderberg, K.~Simonyan, A.~Zisserman, et~al.
\newblock Spatial transformer networks.
\newblock In {\em NIPS}, 2015.

\bibitem{Jia2014caffe}
Y.~Jia, E.~Shelhamer, J.~Donahue, S.~Karayev, J.~Long, R.~Girshick,
  S.~Guadarrama, and T.~Darrell.
\newblock Caffe: Convolutional architecture for fast feature embedding.
\newblock {\em arXiv preprint arXiv:1408.5093}, 2014.

\bibitem{Jiang2016face}
H.~Jiang and E.~G. Learned{-}Miller.
\newblock Face detection with the faster {R-CNN}.
\newblock {\em CoRR}, abs/1606.03473, 2016.

\bibitem{Dai2016instance}
J.~S. Jifeng~Dai, Kaiming~He.
\newblock Instance-aware semantic segmentation via multi-task network cascades.
\newblock In {\em CVPR}, 2016.

\bibitem{Jin2015aligning}
J.~Jin, K.~Fu, R.~Cui, F.~Sha, and C.~Zhang.
\newblock Aligning where to see and what to tell: image caption with
  region-based attention and scene factorization.
\newblock {\em CoRR}, abs/1506.06272, 2015.

\bibitem{Johnson2015densecap}
J.~Johnson, A.~Karpathy, and L.~Fei-Fei.
\newblock Densecap: Fully convolutional localization networks for dense
  captioning.
\newblock {\em CVPR}, 2016.

\bibitem{Karpathy2015deep}
A.~Karpathy and L.~Fei-Fei.
\newblock Deep visual-semantic alignments for generating image descriptions.
\newblock In {\em CVPR}, 2015.

\bibitem{Karpathy2014deep}
A.~Karpathy, A.~Joulin, and L.~Fei~Fei.
\newblock Deep fragment embeddings for bidirectional image sentence mapping.
\newblock In {\em NIPS}, 2014.

\bibitem{Krishna2016visual}
R.~Krishna, Y.~Zhu, O.~Groth, J.~Johnson, K.~Hata, J.~Kravitz, S.~Chen,
  Y.~Kalantidis, L.-J. Li, D.~A. Shamma, et~al.
\newblock Visual genome: Connecting language and vision using crowdsourced
  dense image annotations.
\newblock {\em arXiv preprint arXiv:1602.07332}, 2016.

\bibitem{Lecun1998gradient}
Y.~LeCun, L.~Bottou, Y.~Bengio, and P.~Haffner.
\newblock Gradient-based learning applied to document recognition.
\newblock {\em Proceedings of the IEEE}, 86(11):2278--2324, 1998.

\bibitem{Li2017ViPCNN}
Y.~Li, W.~Ouyang, and X.~Wang.
\newblock Vip-cnn: A visual phrase reasoning convolutional neural network for
  visual relationship detection.
\newblock {\em CoRR}, abs/1702.07191, 2017.

\bibitem{Liu2015ssd}
W.~Liu, D.~Anguelov, D.~Erhan, C.~Szegedy, and S.~Reed.
\newblock Ssd: Single shot multibox detector.
\newblock {\em arXiv preprint arXiv:1512.02325}, 2015.

\bibitem{Long2015fully}
J.~Long, E.~Shelhamer, and T.~Darrell.
\newblock Fully convolutional networks for semantic segmentation.
\newblock In {\em CVPR}, 2015.

\bibitem{Lu2016visual}
C.~Lu, R.~Krishna, M.~Bernstein, and L.~Fei-Fei.
\newblock Visual relationship detection with language priors.
\newblock In {\em ECCV}, 2016.

\bibitem{Malinowski2015ask}
M.~Malinowski, M.~Rohrbach, and M.~Fritz.
\newblock Ask your neurons: A neural-based approach to answering questions
  about images.
\newblock In {\em ICCV}, 2015.

\bibitem{Mao2016unambiguous}
J.~Mao, J.~Huang, A.~Toshev, O.~Camburu, A.~L. Yuille, and K.~Murphy.
\newblock Generation and comprehension of unambiguous object descriptions.
\newblock In {\em CVPR}, 2016.

\bibitem{Mao2014explain}
J.~Mao, W.~Xu, Y.~Yang, J.~Wang, and A.~L. Yuille.
\newblock Explain images with multimodal recurrent neural networks.
\newblock {\em arXiv preprint arXiv:1410.1090}, 2014.

\bibitem{Mikolov2010recurrent}
T.~Mikolov, M.~Karafi{\'a}t, L.~Burget, J.~Cernock{\`y}, and S.~Khudanpur.
\newblock Recurrent neural network based language model.
\newblock In {\em Interspeech}, 2010.

\bibitem{Mottaghi2014role}
R.~Mottaghi, X.~Chen, X.~Liu, N.-G. Cho, S.-W. Lee, S.~Fidler, R.~Urtasun, and
  A.~Yuille.
\newblock The role of context for object detection and semantic segmentation in
  the wild.
\newblock In {\em CVPR}, 2014.

\bibitem{Nagaraja2016refer}
V.~K. Nagaraja, V.~I. Morariu, and L.~S. Davis.
\newblock Modeling context between objects for referring expression
  understanding.
\newblock In {\em ECCV}, 2016.

\bibitem{Papineni2002bleu}
K.~Papineni, S.~Roukos, T.~Ward, and W.-J. Zhu.
\newblock Bleu: a method for automatic evaluation of machine translation.
\newblock In {\em ACL}, 2002.

\bibitem{Redmon2015you}
J.~Redmon, S.~Divvala, R.~Girshick, and A.~Farhadi.
\newblock You only look once: Unified, real-time object detection.
\newblock {\em CVPR}, 2016.

\bibitem{Ren2015faster}
S.~Ren, K.~He, R.~Girshick, and J.~Sun.
\newblock Faster r-cnn: Towards real-time object detection with region proposal
  networks.
\newblock In {\em NIPS}, 2015.

\bibitem{Simonyan2014very}
K.~Simonyan and A.~Zisserman.
\newblock Very deep convolutional networks for large-scale image recognition.
\newblock {\em arXiv preprint arXiv:1409.1556}, 2014.

\bibitem{Sutskever2011generating}
I.~Sutskever, J.~Martens, and G.~E. Hinton.
\newblock Generating text with recurrent neural networks.
\newblock In {\em ICML}, 2011.

\bibitem{Vedantam2015cider}
R.~Vedantam, C.~Lawrence~Zitnick, and D.~Parikh.
\newblock Cider: Consensus-based image description evaluation.
\newblock In {\em CVPR}, 2015.

\bibitem{Vinyals2015show}
O.~Vinyals, A.~Toshev, S.~Bengio, and D.~Erhan.
\newblock Show and tell: A neural image caption generator.
\newblock In {\em ICCV}, 2015.

\bibitem{Werbos1988generalization}
P.~J. Werbos.
\newblock Generalization of backpropagation with application to a recurrent gas
  market model.
\newblock {\em Neural Networks}, 1(4):339--356, 1988.

\bibitem{Xu2015show}
K.~Xu, J.~Ba, R.~Kiros, K.~Cho, A.~Courville, R.~Salakhutdinov, R.~S. Zemel,
  and Y.~Bengio.
\newblock Show, attend and tell: Neural image caption generation with visual
  attention.
\newblock In {\em NIPS}, 2015.

\bibitem{Young2014image}
P.~Young, A.~Lai, M.~Hodosh, and J.~Hockenmaier.
\newblock From image descriptions to visual denotations: New similarity metrics
  for semantic inference over event descriptions.
\newblock {\em TACL}, 2:67--78, 2014.

\bibitem{Yu2016refer}
L.~Yu, P.~Poirson, S.~Yang, A.~C. Berg, and T.~L. Berg.
\newblock Modeling context in referring expressions.
\newblock In {\em ECCV}, 2016.

\end{thebibliography}
}

\end{document}